# Duelist Algorithm: An Algorithm Inspired by How Duelist Improve Their Capabilities in a Duel

Totok Ruki Biyanto*[a], Henokh Yernias Fibrianto[a], Gunawan Nugroho[a],
Erny Listijorini[b], Titik Budiati[c], Hairul Huda[d]
[a]Engineering Physics Department, Institut Teknologi Sepuluh Nopember (ITS), Surabaya, Indonesia
[b]Mechanical Engineering Department, Universitas Sultan Ageng Tirtayasa, Cilegon, Indonesia
[c]Food Technology Department, State Polytechnic of Jember, Indonesia
[d]Chemical Engineering Department, Universitas Mulawarman, Samarinda, Indonesia

Highlight:
- A new optimization algorithm inspired by how duelist improve their skill in duel
- In duelist algorithm, different treatment is given to each duelist based on the duel result
- Duelist algorithm provided good result compared to the other optimization algorithms such as genetic algorithm, particle swarm optimization algorithm and imperialist competitive algorithm

*Abstract*—This paper proposes an optimization algorithm based on how human fight and learn from each duelist. Since this algorithm is based on population, the proposed algorithm starts with an initial set of duelists. The duel is to determine the winner and loser. The loser learns from the winner, while the winner try their new skill or technique that may improve their fighting capabilities. A few duelists with highest fighting capabilities are called as champion. The champion train a new duelists such as their capabilities. The new duelist will join the tournament as a representative of each champion. All duelist are re-evaluated, and the duelists with worst fighting capabilities is eliminated to maintain the amount of duelists. Two optimization problem is applied for the proposed algorithm, together with genetic algorithm, particle swarm optimization and imperialist competitive algorithm. The results show that the proposed algorithm is able to find the better global optimum and faster iteration.

## 1. Introduction

Optimization is a process to achieve something better. For example, let there be a problem f(x) then optimization is a process to find optimum value of x which is can be maximum, minimum or at specific value in between. Different methods have been proposed to solve the optimization problem. One of the methods is genetic algorithm (GA) which is based on natural selection by evolving a population of candidate solution for defined objective function [1]. On the other hand, a different method for optimization called ant colony optimization is inspired by foraging behavior of real ants [2]. Another type of method is inspired by social behavior of animals which is called as particle swarm optimization (PSO) [3]. There's also a method for optimization which inspired by imperialistic competition called imperialist competitive algorithm (ICA) [4]. All of these mentioned methods are population based algorithm which is mean that there's a set of population and keep improving itself in each iterations [5]. There are other optimization algorithms also commonly used such as predatory search strategy [6], optimization algorithm based on bacterial chemotaxis [7], bacterial foraging optimization [8], society and civilization optimization [9], group search optimizer [10], chemical reaction optimization [11] and quantum evolutionary algorithm [12]. Nowadays, all this optimization methods are very useful for solving multiple problems starting from process industry, energy management, scheduling, resource allocation or even pattern recognition and machine learning [13-20].

In this paper, a new algorithm based on genetic algorithm is proposed which is inspired by human fighting and learning capabilities. As an overview, in genetic algorithm there are two ways to develop an individual into a new one. First is crossover where an individual mate with another individual to produce a new offspring, this new offspring's genotype are based on their parents. The second one is mutation where an individual mutate into a new one. In duelist algorithm (DA), all the individual in population are called as duelist, all those duelists would fight one by one to determine the champions, winners and losers. The fighting itself just like real life fight where the strongest has posiblilty as a loser. There is a probability that the weak one would be lucky enough to win. In order to improve each duelist, there are also two ways to evolve. One of them is innovation which is similar to mutation in genetic algorithm. The difference is only the winners would possibly to be innovated. The other one is called as learning, losers would learn from winners. In genetic algorithm, both mutation and crossover are seem to be blind in producing any solution to find the best solution. Blind means that each solution or produced individu in genetic algorithm may has not better solution. In fact, it may fall into the worst one. Duelist algorithm tries to minimize this blind effect by giving different treatment on each duelist based on their classification. This paper described how duelist algorithm is designed and implemented.

―――――
*Corresponding author.
  E-mail address: trb@ep.its.ac.id, trbiyanto@mail.com (Totok R. Biyanto)

## 2. Review of A Duel

Duel can be interpreted as a fighting between one or more person(s) with other person(s). Fighting require physical strength, skill and intellectual capability, for example in chess and bridge games. Common type of duel which include physical strength is boxing, boxing is one of world's most popular sport where two persons need to knock down each of them under certain rules [21]. Soccer is also categorized as a duel where two teams must have higher score goals to win the match, soccer is much more complicated than boxing strategy that teamwork plays important role [22]. In every duel, there are consist of the winner and the loser as well as the rules. Take soccer for example, winner in soccer match is defined as a team which has more goal score than their opponent and each team must be obey the rules. In a match the probability become the winner depend on strength, skill and lucky. After the match, knowing the capabilities of the winner and the loser are very useful. Loser can learn from how the winner, and winner can improve the capability and skill by training or trying something new from the loser. In the proposed algorithm, each duelist do the same to be unbeatable, by upgrading themselves whether by learning from their opponent or developing a new technique or skill.

## 3. Duelist Algorithm

The flowchart of proposed algorithm is shown in Figure 1. First, population of duelist is registered. Each duelist has their properties which is encoded into binary array. Each duelist is evaluated to determine their fighting capabilities. The duel schedule is set to each duelist that contain a set of duel participants. In the duel, each duelist would fight one on one with other duelist. This one on one fighting is used rather than gladiator battle to avoid local optimum. Each duel would produce a winner and a loser based on their fighting capabilities and their 'luck'. After the match, the champion is also determined. This champions are the duelist that has the best fighting capabilities.

In the next step, each winner and loser have opportunity to upgrade their fighting capabilities, meanwhile each champion train the new duelist as such their capabilitis. The new duelist will join in the next match. Each loser would learn from their opponents how to be a better duelist by replacing a specific part of their binary array with winner's binary array value. On the other hand, winner would try to innovate a new technique or skill by changing their binary array value into something new.

Each duelist fighting capabilities is re-evaluated for the next match. All the duelists then re-evaluated through post-qualification and sorted to determine who deserve to be champions. Since there are new duelists that was trained by champions, all the worst duelists are eliminated to maintain the amount of duelists in the tournament. Classification of each duelist is shown in Figure 2. This process will continue until the tournament is finished. The step by step explnation as follow:

*A. Registration of Duelist Candidate*

Each duelist in a duelist set is registered using binary array. Binary array in genetic algorithm is called as chromosome, however in duelist algorithm is called as skillset. In a $N_{var}$-dimensional optimization problem, the duelist would be binary length times $N_{var}$ length array.

*B. Pre-Qualification*

Pre-qualification is a test that given to each duelists to measure or evaluate their fighting capabilities based on their skillset.

*C. Determination of Champions Board*

Board of champions is determined to keep the best duelist in the game. Each champion should trains a new duelist to be as well as himself duel capabilities. This new duelists would replace the champion position in the game and join the next duel.

*D. Define A Duel Schedule Between Each Duelist*

The duel schedule between each duelist is set randomly. Each duelist will fight using their fighting capabilities and luck to determine the winner and the loser. The duel is using a simple logic. If duelist A's fighting capabilities plus his luck are higher than duelist B's, then duelist A is the winner and vice versa. Duelist's luck is detemined by purely random function to avoid local optimum. The pseudocode to determine the winner and the loser is shown in Algorithm 1.

Algorithm 1. Determination of the winner and the loser.

```
Require : Duelist A and B; Luck_Coefficient
A(Luck) = A(Fighting_Capabilities) * (Luck_Coefficient + (rand(0-1) * Luck_Coefficient));
B(Luck) = B(Fighting_Capabilities) * (Luck_Coefficient + (rand(0-1) * Luck_Coefficient));
If  ((A(Fighting_Capabilities) + A(Luck)) <= (B(Fighting_Capabilities) + B(Luck)))
    A(Winner) = 1;
    B(Winner) = 0;
Else
    A(Winner) = 0;
    B(Winner) = 1;
End
```

*E. Duelist's Improvement*

After the match, each duelist are categorized into champion, winner and loser. To improve each duelist fighting capabilities there are three kind of treatment for each catagories. First treatment is for losers, each loser is trained by learning from winner. Learning means that loser may copy a part of winner's skillset or binary array. The second treatment is for winners, each winner would improve their own capabilities by trying some thing new from the loser. This treatment consist of winner's binary array random manipulation. And the last, each champion would trains a new duelist.

*F. Elimination*

Since there are some new duelists joining the game, there must be an elimination to keep duelists quantity still the same as defined before. Elimination is based on each duelist's dueling capabilities. The duelist with worst dueling capabilities are eliminated.

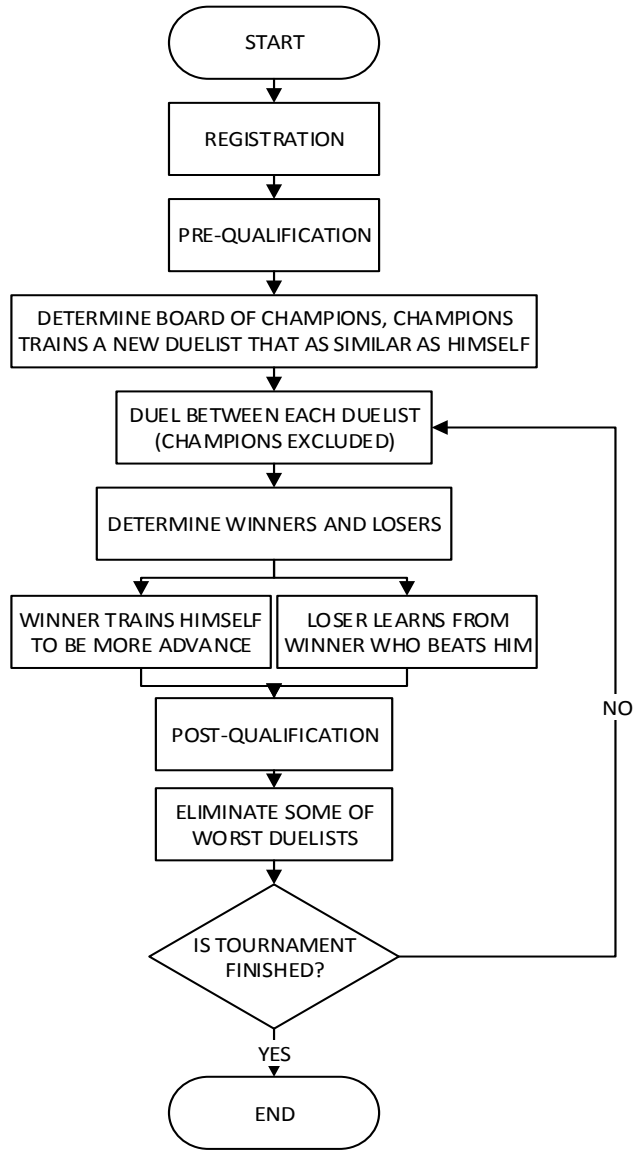

Fig. 1. Duelist Algorithm flowchart

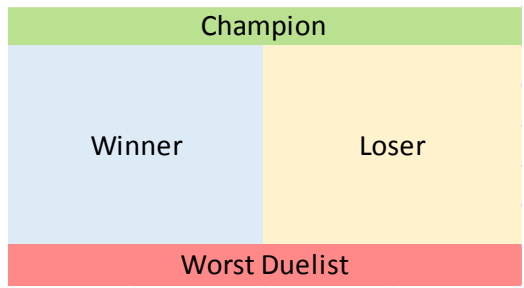

Fig. 2. Duelist's Classification.

## 4. Experimental Studies

This section discuss about Duelist Algorithm performance using 2 benchmarks. All these problems are maximization problems. The detail of these functions are shown as follow:

Problem $M_1$ :

$$f = -(x.\sin(4.x) + 1{,}1.y.\sin(2.y)) \tag{1}$$

Problem $M_2$ :

$$f = -\left(\sqrt{x^2+y^2}.\cos(x-y).e^{\left(\cos(\frac{x.(y+5)}{7})\right)}\right) \quad (2)$$

Shifted Sphere function :

$$f = \sum_{i=1}^{D} x^2 + f\_bias \quad (3)$$

Figure 3 shows a 3D plot of function of problem $M_1$ in interval $0 < x,y < 10$.

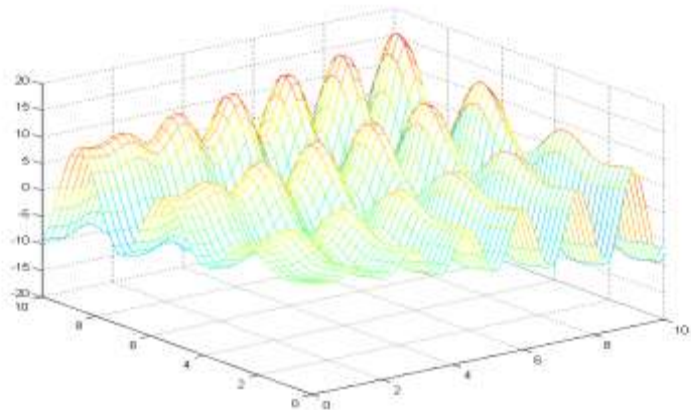

Fig. 3. 3D plot of function in problem $M_1$

The initial population of 100 duelists, max generations of 200, luck coefficient of 0 and mutation probability of 0.5 is set. Result of first test is shown in Figure 4.

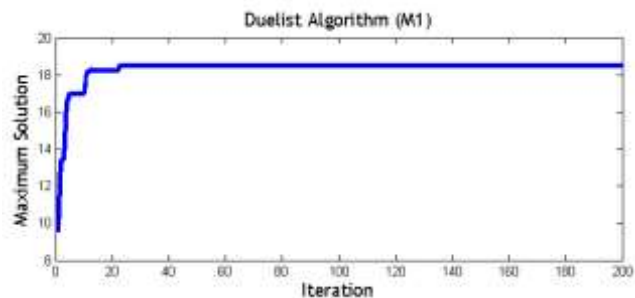

Fig. 4. Duelist Algorithm's maximum cost versus iteration in problem $M_1$.

Maximum value is found at 18.5285 which is supposed to be 18.5547. For comparison, a GA is applied with population of 100 individuals, max generations 200, mutation probability of 0.5 and crossover probability of 0.8. To eliminate any unfair factor such as random population, predetermined population is used for both algorithm. Since both of them have a pseudo-random function (e.g. mutation), both of them always provide different result every test. Figure 5 shows GA and DA comparison for the first test using $M_1$ as optimization problem.

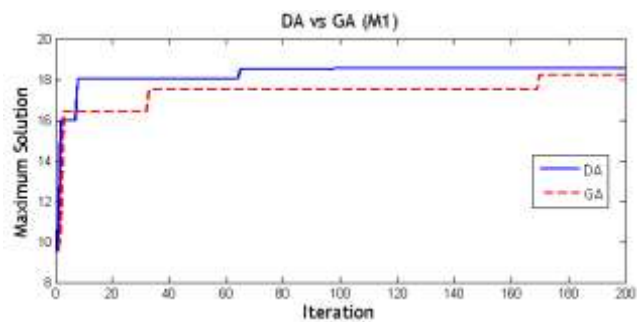

Fig. 5. Duelist Algorithm and Genetic Algorithm's maximum cost versus iteration in problem $M_1$.

Another type of problem called $M_2$ is also applied to GA and duelist algorithm. Figure 6 shows 3D plot function of problem $M_2$ in interval $0 < x,y < 10$ with maximum value 30.3489.

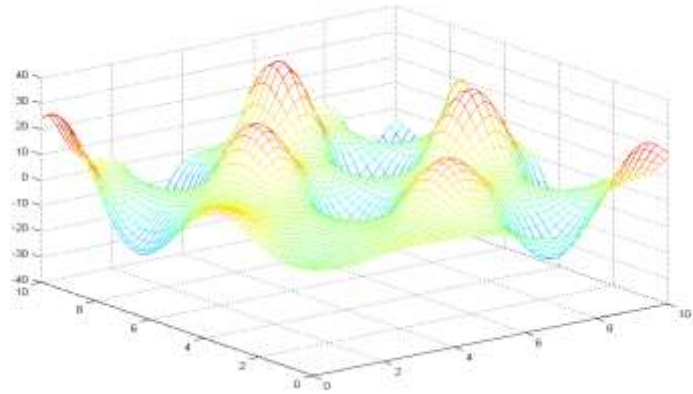

Fig. 6. 3D plot of function in problem $M_2$

Test is perform using same parameter, and the result is shown in Figure 7. Figure 7 shows that Duelist Algorithm achieve faster and better solution which is in 143 iterations and maximum value at 30.3060. While GA spends 166 iterations to reach the maximum solution and find the maximum value at 30.3017. To provide a fair comparison, 10 tests of both algorithm for problem M2 is performed in different optimization parameters, and the results are shown at Figure 8 and Figure 9.

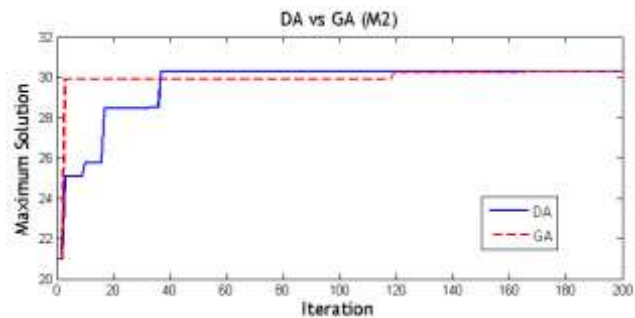

Fig. 7. Duelist Algorithm and Genetic Algorithm's maximum cost versus iteration in problem $M_2$.

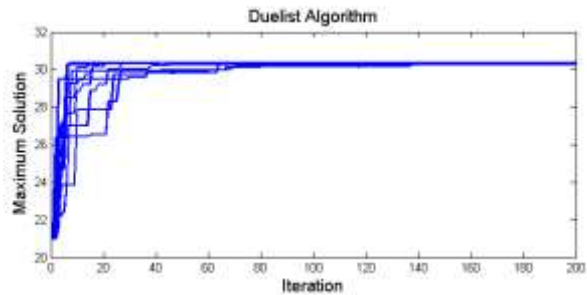

Fig. 8. Duelist Algorithm's maximum solutions versus iteration in problem $M_2$.

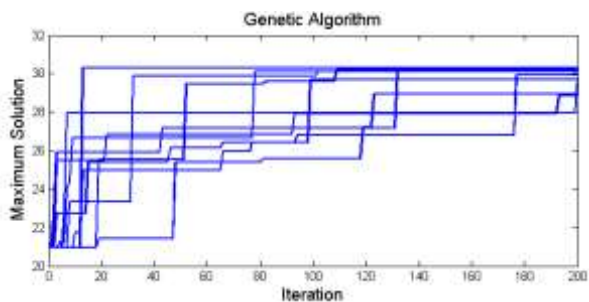

Fig. 9. Genetic Algorithm's maximum solutions versus iteration in problem $M_2$.

Particle Swarm Optimization (PSO) and Imperialist Competitive Algorithm (ICA) is also utilized for other comparisons. Figure 10 shows the result of ICA using 100 countries, revolution rate of 0.3, assimilation coefficient of 2, assimilation angle coefficient of 0.5, zeta of 0.02, damp ration of 0.99, 8 initial imperialists and 200 decades. Figure 11 shows the result of PSO using 100 swarms, 200 iterations, speed constants of 0.4 and 0.6, and theta within range of 0.5 to 0.9. The experiments show that DA is faster than PSO and GA in reaching global optimum.

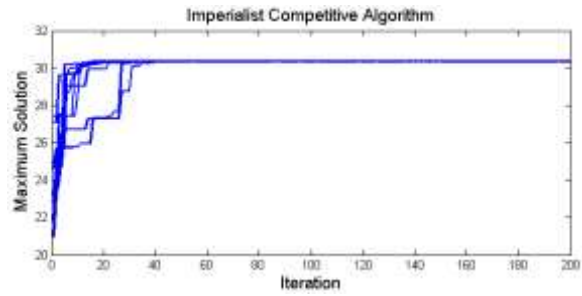

Fig. 10. ICA's maximum solutions versus iteration in problem $M_2$.

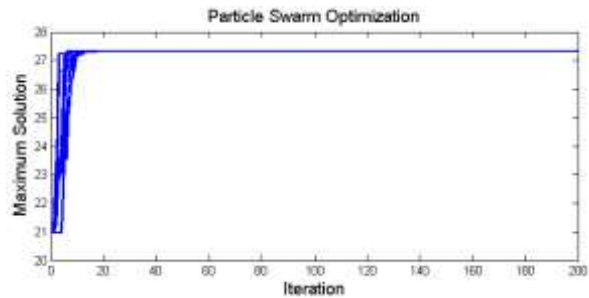

Fig. 11. PSO's maximum solutions versus iteration in problem $M_2$.

To evaluate the impact of each parameter in duelist algorithm, a series of test was taken by changing in the innovation probability, learning probability and luck coefficient. In this test, shifted sphere function is used to evaluate duelist algorithm performance based on its parameter. Table 1 shows the same best solution is that obtained at different learning probability, however the best solution is achieved at different iteration. The fastest iteration to achieved the best solution is provided by learning probability 0.2, while the innovation probability and luck coefficient are 0.1 and 0.01, respectively.

Table 1. Variation of the best solution by change in the learning probability

| Learning probability | 0.1 | 0.2 | 0.3 | 0.4 | 0.5 | 0.6 | 0.7 | .0.8 | 0.9 | 1.0 |
|---|---|---|---|---|---|---|---|---|---|---|
| Best solution | 449.997 | 449.997 | 449.997 | 449.997 | 449.997 | 449.997 | 449.997 | 449.997 | 449.997 | 449.997 |
| Iteration | 12 | 4 | 7 | 9 | 15 | 15 | 6 | 9 | 14 | 10 |

In the same way, The result shows that duelist algorithm provide the fastest iteration to achieved the best solution is obtained by using innovation probability 0.1, while the learning probability and luck coefficient are 0.2 and 0.01, respectively, as shown in Table 2. Table 3 shows the fastest iteration to achieved the best solution is provided by luck coefficient 0.25, while the learning probability and innovation probability are 0.2 and 0.4, respectively.

Table 2. Variation of best solution by change in innovation probability

| Innovation probability | 0.1 | 0.2 | 0.3 | 0.4 | 0.5 | 0.6 | 0.7 | .0.8 | 0.9 | 1.0 |
|---|---|---|---|---|---|---|---|---|---|---|
| Best solution | 449.997 | 449.997 | 449.997 | 449.997 | 449.997 | 449.997 | 449.997 | 449.997 | 449.997 | 449.802 |
| Iteration | 4 | 8 | 9 | 5 | 8 | 27 | 26 | 29 | 13 | 14 |

Table 3. Variation of best solution with change in luck coefficient

| Luck coefficient | 0.05 | 0.10 | 0.15 | 0.20 | 0.25 | 0.30 | 0.35 | .0.40 | 0.45 | 0.50 |
|---|---|---|---|---|---|---|---|---|---|---|
| Best solution | 449.997 | 449.997 | 449.997 | 449.997 | 449.997 | 449.997 | 449.997 | 449.997 | 449.997 | 449.997 |
| Iteration | 7 | 16 | 10 | 15 | 4 | 19 | 20 | 13 | 11 | 11 |

In general, the best solution is achived with different iterations, it depend on learning probability, innovation probability and luck coefficient as optimization parameter. The different iteration time for each different parameters are less tha a second. Hence DA as a new stochastic optimization may be utilized as an alternatif optimization technique to seek global optimum solution.

## 5. Conclusion

In this paper, an optimization algorithm based on how duelist improve himself to win a fight is proposed. Each individual in the population is called duelist. Each duelist fight with other duelist to determine who is the winner and the loser. Winner and loser have their own way to improve theirself. The winners are improved by learning theirself. On the other hand, the loser improve himself by learning from the winner. After several improvements and duels, some duelists will become the best solution for given problem. The algorithm is tested by using 2 optimization problems or benchmarks. The results shows that the proposed algorithm is able to find the better global optimum with faster iteration compared to genetic algorithm, particle swarm optimization and imperialist competitive algorithm.

**Acknowledgment**

The authors gratefully thank to Sepuluh Nopember Institute of Techology (ITS) Surabaya, Universitas Sultan Ageng Tirtayasa, Universitas Mulawarman and State Polytechnic of Jember for providing the facilities for conducting this research.

## References

1. Mitchell M (1998) An introduction to genetic algorithms. MIT press,
2. Dorigo M, Blum C (2005) Ant colony optimization theory: A survey. Theoretical computer science 344 (2):243-278
3. Poli R, Kennedy J, Blackwell T (2007) Particle swarm optimization. Swarm intelligence 1 (1):33-57
4. Atashpaz-Gargari E, Lucas C Imperialist competitive algorithm: an algorithm for optimization inspired by imperialistic competition. In: Evolutionary computation, 2007. CEC 2007. IEEE Congress on, 2007. IEEE, pp 4661-4667
5. Beasley JE, Chu PC (1996) A genetic algorithm for the set covering problem. European Journal of Operational Research 94 (2):392-404
6. Linhares A (1999) Synthesizing a predatory search strategy for VLSI layouts. Evolutionary Computation, IEEE Transactions on 3 (2):147-152
7. Müller SD, Marchetto J, Airaghi S, Kournoutsakos P (2002) Optimization based on bacterial chemotaxis. Evolutionary Computation, IEEE Transactions on 6 (1):16-29
8. Passino KM (2002) Biomimicry of bacterial foraging for distributed optimization and control. Control Systems, IEEE 22 (3):52-67
9. Ray T, Liew KM (2003) Society and civilization: An optimization algorithm based on the simulation of social behavior. Evolutionary Computation, IEEE Transactions on 7 (4):386-396
10. He S, Wu QH, Saunders J (2009) Group search optimizer: an optimization algorithm inspired by animal searching behavior. Evolutionary Computation, IEEE Transactions on 13 (5):973-990
11. Lam A, Li VO (2010) Chemical-reaction-inspired metaheuristic for optimization. Evolutionary Computation, IEEE Transactions on 14 (3):381-399
12. Han K-H, Kim J-H (2002) Quantum-inspired evolutionary algorithm for a class of combinatorial optimization. Evolutionary Computation, IEEE Transactions on 6 (6):580-593
13. Hartmann S (1998) A competitive genetic algorithm for resource-constrained project scheduling. Naval Research Logistics (NRL) 45 (7):733-750
14. Dandy GC, Simpson AR, Murphy LJ (1996) An improved genetic algorithm for pipe network optimization. Water Resources Research 32 (2):449-458
15. Jones G, Willett P, Glen RC (1995) Molecular recognition of receptor sites using a genetic algorithm with a description of desolvation. Journal of molecular biology 245 (1):43-53
16. Johnston RL, Cartwright HM (2004) Applications of evolutionary computation in chemistry, vol 110. Springer Science & Business Media,
17. Deb K, Pratap A, Agarwal S, Meyarivan T (2002) A fast and elitist multiobjective genetic algorithm: NSGA-II. Evolutionary Computation, IEEE Transactions on 6 (2):182-197
18. Balci HH, Valenzuela JF (2004) Scheduling electric power generators using particle swarm optimization combined with the Lagrangian relaxation method. International Journal of Applied Mathematics and Computer Science 14 (3):411-422
19. Colombetti M, Dorigo M (1999) Evolutionary computation in behavior engineering. Evolutionary computation: theory and applications:37-80
20. Fogel DB (1988) An evolutionary approach to the traveling salesman problem. Biological Cybernetics 60 (2):139-144
21. Woodward K (2006) Boxing, Masculinity and Identity. The" I" of the Tiger. Routledge.
22. Reilly T (2003) Science and soccer. Routledge.